# Portfolio selection using neural networks

Alberto Fernández, Sergio Gómez*

*Departament d'Enginyeria Informàtica i Matemàtiques, Universitat Rovira i Virgili, Campus Sescelades, Avinguda dels Països Catalans 26, E-43007 Tarragona, Spain*

**Abstract**

In this paper we apply a heuristic method based on artificial neural networks in order to trace out the efficient frontier associated to the portfolio selection problem. We consider a generalization of the standard Markowitz mean-variance model which includes cardinality and bounding constraints. These constraints ensure the investment in a given number of different assets and limit the amount of capital to be invested in each asset. We present some experimental results obtained with the neural network heuristic and we compare them to those obtained with three previous heuristic methods.

**Scope and purpose**

The portfolio selection problem is an instance from the family of quadratic programming problems when the standard Markowitz mean-variance model is considered. But if this model is generalized to include cardinality and bounding constraints, then the portfolio selection problem becomes a mixed quadratic and integer programming problem. When considering the latter model, there is not any exact algorithm able to solve the portfolio selection problem in an efficient way. The use of heuristic algorithms in this case is imperative. In the past some heuristic methods based mainly on evolutionary algorithms, tabu search and simulated annealing have been developed. The purpose of this paper is to consider a particular neural network model, the Hopfield network, which has been used to solve some other optimisation problems and apply it here to the portfolio selection problem, comparing the new results to those obtained with previous heuristic algorithms.

*Keywords:* Portfolio selection; Efficient frontier; Neural networks; Hopfield network

## 1. Introduction

In the portfolio selection problem, given a set of available securities or assets, we want to find out the optimum way of investing a particular amount of money in these assets. Each one of the different ways to diversify this money between the several assets is called a portfolio. For solving this portfolio selection problem Markowitz [1] presented the so called mean-variance model, which assumes that the total return of a portfolio can be described using the mean return of the assets and the variance of return (risk) between these assets. The portfolios that offer the minimum risk for a given level of return form what it is called the efficient frontier. For every level of desired mean return, this efficient frontier gives us the best way of investing our money.

However, the standard mean-variance model has not got any cardinality constraint ensuring that every portfolio invests in a given number of different assets, neither uses any bounding constraint limiting the amount of money to be invested in each asset. This sort of constraints are very useful in practice. In order to overcome these inconveniences, the standard model can be generalized to include these constraints.

In this paper we focus on the problem of tracing out the efficient frontier for the general mean-variance model with cardinality and bounding constraints. In previous work, some heuristic methods have been developed for the portfolio selection problem. There are some methods that use evolutionary algorithms [2-6], tabu search [2,7] and simulated annealing [2,8,9]. Here we present a different heuristic method based on artificial neural networks. The results obtained are compared to those obtained using three representative methods from [2] based on genetic algorithms, tabu search and simulated annealing.

* Corresponding author. Tel.: +34 977 55 8508; fax: +34 977 55 9710.
  *E-mail address:* sergio.gomez@urv.net (S. Gómez)

Following this introduction, in the second section we present the model formulation for the portfolio selection problem. The third section describes the Hopfield neural network as well as the way to use it for solving this problem. In the fourth section we present some experimental results and, in the fifth section, we finish with some conclusions.

## 2. Portfolio selection

First of all, as we introduce the notation that we are going to use in this paper, let us remember the well known Markowitz mean-variance model [1] for the portfolio selection problem. Let $N$ be the number of different assets, $\mu_i$ be the mean return of asset $i$, $\sigma_{ij}$ be the covariance between returns of assets $i$ and $j$, and let $\lambda \in [0,1]$ be the risk aversion parameter. The decision variables $x_i$ represent the proportion of capital to be invested in asset $i$. Using this notation, the standard Markowitz mean-variance model for the portfolio selection problem is:

$$\text{minimise} \quad \lambda \left[ \sum_{i=1}^{N} \sum_{j=1}^{N} x_i \sigma_{ij} x_j \right] + (1-\lambda) \left[ -\sum_{i=1}^{N} \mu_i x_i \right] \quad (1)$$

$$\text{subject to} \quad \sum_{i=1}^{N} x_i = 1 \quad (2)$$

$$0 \leq x_i \leq 1, \quad i = 1,...,N \quad (3)$$

The case with $\lambda=0$ represents maximising the portfolio mean return (without considering the variance) and the optimal solution will be only formed by the asset with the greatest mean return. The case with $\lambda=1$ represents minimising the total variance associated to the portfolio (regardless of the mean returns) and the optimal solution will typically consist of several assets. Any value of $\lambda$ inside the interval (0,1) represents a tradeoff between mean return and variance, generating a solution between the two extremes $\lambda=0$ and $\lambda=1$.

Since every solution satisfying all the constraints (feasible solution) corresponds with one of the possible portfolios, from here on we will speak without distinguishing between solutions for the above problem and portfolios.

The portfolio selection problem is an instance of the family of multiobjective optimisation problems. So, one of the first things to do is to adopt a definition for the concept of optimal solution. Here we will use the Pareto optimality definition [10]. That is, a feasible solution of the portfolio selection problem will be an optimal solution (or non dominated solution) if there is not any other feasible solution improving one objective without making worse the other.

Usually a multiobjective optimisation problem has several different optimal solutions. The objective function values of all these non dominated solutions form what it is called the efficient frontier. For the problem defined in Eqs. (1)-(3), the efficient frontier is an increasing curve that gives the best tradeoff between mean return and variance (risk). In Fig. 1 we show an example of such a curve corresponding to the test data described in Section 4. This efficient frontier has been computed taking 2000 different values for the risk aversion parameter $\lambda$ and solving exactly the corresponding portfolio selection problems. The objective function values of the resulting solutions give the 2000 points that form the curve in Fig. 1. We call this curve the standard efficient frontier in order to distinguish it from the general efficient frontier, corresponding to the general mean-variance portfolio selection model.

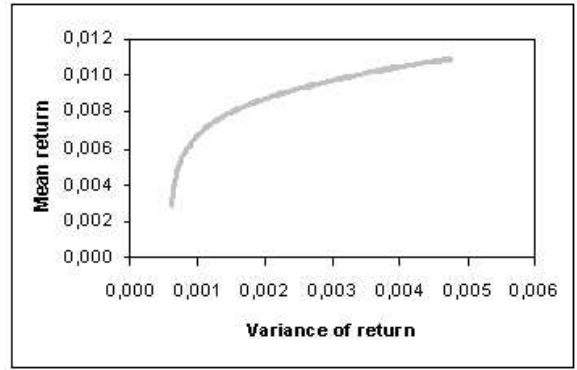

Fig. 1. Standard efficient frontier.

With the purpose of generalizing the standard Markowitz model to include cardinality and bounding constraints, we will use a model formulation that can be also found in [2,7,11]. In addition to the previously defined variables, let $K$ be the desired number of different assets in the portfolio with no null investment, $\varepsilon_i$ and $\delta_i$ be respectively the lower and upper bounds for the proportion of capital to be invested in asset $i$, with $0 \leq \varepsilon_i \leq \delta_i \leq 1$. The additional decision variables $z_i$ are 1 if asset $i$ is included in the portfolio and 0 otherwise. The general mean-variance model for the portfolio selection problem is:

$$\text{minimise} \quad \lambda \left[ \sum_{i=1}^{N} \sum_{j=1}^{N} x_i \sigma_{ij} x_j \right] + (1-\lambda) \left[ -\sum_{i=1}^{N} \mu_i x_i \right] \quad (4)$$

$$\text{subject to} \quad \sum_{i=1}^{N} x_i = 1 \quad (5)$$

$$\sum_{i=1}^{N} z_i = K \quad (6)$$

$$\varepsilon_i z_i \leq x_i \leq \delta_i z_i, \qquad i = 1,...,N \quad (7)$$

$$z_i \in \{0,1\}, \qquad i = 1,...,N \quad (8)$$

This formulation is a mixed quadratic and integer programming problem for which efficient algorithms do no exist. Another difference with the standard model is that in the presence of cardinality and bounding constraints the resulting efficient frontier, which we are going to call general efficient frontier, can be quite different from the one obtained with the standard mean-variance model. In particular, the general efficient frontier may be discontinuous [2,11].

## 3. The Hopfield network

### 3.1. Description

There are two main approaches for solving combinatorial optimisation problems using artificial neural networks: Hopfield networks and Kohonen's self-organizing feature maps. While the latter are mainly used in Euclidean problems, the Hopfield networks have been widely applied in different classes of combinatorial optimisation problems [12]. Although the problem at hand is not a combinatorial optimisation one, we take advantage of the fact that the objective function in Eq. (4) has the same form than the energy function in Hopfield networks and, consequently, it will be minimised if we follow the Hopfield dynamics.

The Hopfield network [13] is an artificial neural network model with a single layer of neurons fully connected. That is, all of the $N$ neurons in the network are connected to each other as well as to themselves. The equations (with discrete time) that govern the dynamics of this network are:

$$x_i(t+1) = G_i\left(\sum_{j=1}^{N} w_{ji} x_j(t) + b_i\right) \qquad (9)$$

where $x_i(t) \in [\varepsilon, \delta]$ is the state of neuron $i$ at time $t$, $b_i$ is the constant extern input (bias) for neuron $i$, and $w_{ji}$ is the weight of the synaptic connection from neuron $j$ to neuron $i$. $G_i: R \rightarrow [\varepsilon, \delta]$ is the activation function and usually has the form of a sigmoid with a gain $\beta_i > 0$:

$$G_i(y) = \varepsilon + \frac{\delta - \varepsilon}{1 + \exp(-\beta_i y)} \qquad (10)$$

The output vector in a Hopfield network represents the solution for the problem at hand. This vector lies inside the hypercube $[\varepsilon, \delta]^N$. The stability of the network can be proved defining a so called energy function for the network and proving that its time derivative is nonincreasing.

The nonlinear nature of the Hopfield network produces multiple equilibrium points. For any given set of initial conditions $x(0)$, the symmetric Hopfield network (with $w_{ji}=w_{ij}$) will converge towards a stable equilibrium point. When the network is deterministic, the position of that point is uniquely determined by the initial conditions: all the initial conditions that lie inside the attraction field of an equilibrium point will converge asymptotically towards that point. The exact number of equilibrium points and their positions are determined by the network parameters $w_{ji}$ and $\beta_i$. When the gain $\beta_i$ is small, the number of equilibrium points is low (possibly as low as 1) and they lie inside the hypercube $[\varepsilon, \delta]^N$. Nevertheless, as the gain increases, the number of equilibrium points also increases and their positions move towards the vertices of the hypercube. When the gain tends to its extreme values ($\beta_i \rightarrow +\infty$), the equilibrium points reach the hypercube vertices and are maximum in number. In this case, the energy function for the network has the following form:

$$E(x) = -\frac{1}{2}\sum_{i=1}^{N}\sum_{j=1}^{N} x_i w_{ij} x_j - \sum_{i=1}^{N} b_i x_i \qquad (11)$$

In this work we update the neurons asynchronously, that is, only one neuron at a time. The neurons to be updated are selected randomly. This way of updating does not change the positions of the equilibrium points in the network, but it does change the descending path through the energy surface. So, initial conditions that originally were attracted to a particular equilibrium point, can be attracted towards a different equilibrium point when using asynchronous updating.

For solving the portfolio optimisation problem, we have implemented a Hopfield network with gains $\beta_i$ changing through time [14]. Initially the gains are very small, producing a single equilibrium point. So, regardless of the initial conditions, the network converges towards that point. Then, as time passes, gains are increased little by little, producing energy surfaces with a higher number of equilibrium points and moving these equilibrium points towards the vertices of the hypercube $[\varepsilon, \delta]^N$.

*3.2. Absence of cycles in symmetric neural networks*

In [15] it is explained that the symmetric Hopfield network ($w_{ji}=w_{ij}$) can converge to a cycle of length 2. The same paper shows that, in order to avoid this undesired behaviour in our network dynamics, the following discrete model can be used:

$$x_i(t+1) = (1-\alpha_i)x_i(t) + \alpha_i G_i\left(\sum_{j=1}^{N} w_{ji} x_j(t) + b_i\right) \qquad (12)$$

with $\alpha_i \in (0,1]$.

When using this discrete model, periodic points that are not fix points can appear, specially when all $\alpha_i=1$. But if synaptic weights are symmetric ($w_{ji}=w_{ij}$) and $w_{ii}>-(2-\alpha_i)/(\alpha_i\beta_i)$, then the above discrete model has the sequential dynamics convergent to fix points for any $\alpha_i \in (0,1]$. Since the synaptic weights $w_{ii}$ are fixed from the beginning and the gains $\beta_i$ are increased little by little, given any particular pair of values $w_{ii}$ and $\beta_i$, what we must do is to give a value to $\alpha_i$ satisfying the previous condition.

*3.3. Energy function for the portfolio selection problem*

Finally we are going to deduce an energy function for the problem we are dealing with and, in doing so, we will get the values for the constants $w_{ji}$ and $b_i$. Let us first remember the objective function of our portfolio selection problem:

$$f(x) = \lambda \left[ \sum_{i=1}^{N} \sum_{j=1}^{N} x_i \sigma_{ij} x_j \right] + (1-\lambda) \left[ -\sum_{i=1}^{N} \mu_i x_i \right] \tag{13}$$

Passing the multiplicative coefficients inside the additions, we have:

$$f(x) = \sum_{i=1}^{N} \sum_{j=1}^{N} x_i (\lambda \sigma_{ij}) x_j - \sum_{i=1}^{N} (1-\lambda) \mu_i x_i \tag{14}$$

For this function to have the same appearance than the energy function in Eq. (11) it is only necessary that the first term appears multiplied by $-1/2$, so we multiply and divide this term by $-2$, getting the following energy function:

$$E(x) = -\frac{1}{2} \sum_{i=1}^{N} \sum_{j=1}^{N} x_i (-2\lambda \sigma_{ij}) x_j - \sum_{i=1}^{N} (1-\lambda) \mu_i x_i \tag{15}$$

By simple comparison of this energy function with the general energy function in Eq. (11), we get the values for the synaptic weights and the extern inputs:

$$w_{ij} = -2\lambda \sigma_{ij} \tag{16}$$
$$b_i = (1-\lambda)\mu_i \tag{17}$$

When solving any optimisation problem using a Hopfield network, the problem constraints usually appear in the energy function. However, in our case it is not necessary.

First, regarding the constraint $x_i \in [\varepsilon_i, \delta_i]$ in Eq. (7), we can say that it will be satisfied using as the threshold function a sigmoid such as the one in Eq. (10), since its outputs already lie inside the desired interval.

To satisfy the cardinality constraint in Eq. (6) we begin our heuristic algorithm with a neural network having $N$ neurons that follow the already explained Hopfield dynamics. In doing so, we get a minimum for the objective function. Next thing to do is pruning the least representative neuron, that is, the one with the smallest output. Then we update this new network (with one less neuron) following the same Hopfield dynamics. These two steps, neuron pruning and objective function minimisation, are repeated until the network has exactly $K$ neurons. These remaining neurons are a solution for our original portfolio selection problem.

We are only left to consider the constraint in Eq. (5). To satisfy this constraint we do the same thing that has been done in [2]: the feasibility of every portfolio is evaluated using a greedy algorithm which changes the proportions of capital $x_i$ to be invested in each selected asset, in order to ensure, if possible, that all constraints are satisfied. In a first step the algorithm assigns to all $x_i$ corresponding to a selected asset its lower limit $\varepsilon_i$ plus a fraction proportional to its current value. This ensures that all the constraints relating to the lower bounds are satisfied. In a second iterative step the algorithm takes all the selected assets exceeding their respective upper limits $\delta_i$ and fixes them up to these upper limits. Then the rest of the selected assets that are not fixed up, are given a new value for $x_i$ ensuring the lower bounds $\varepsilon_i$ and adding a fraction of the free portfolio proportion. This iterative process is repeated until there is not any asset out of its limits.

Bringing together all that we have said until now, in Algorithm 1 we show the neural network heuristic used in this work.

```
function neural_network_heuristic
    Δλ        increment for the risk aversion parameter
    M         number of portfolios in the set P
returns
    H         set with all the Pareto optimal portfolios
var
    P         set of portfolios
    P_min     minimum portfolio
    T         number of iterations
    R         number of repetitions at the same gain value
    P_can     candidate portfolio
begin
    H := ∅
    for λ := 0 to 1 by Δλ do
        P := initialise_portfolios_randomly(M)    {K assets in each one of the M portfolios}
        evaluate_portfolios(P, H)                                          {greedy algorithm}
        P_min := minimum_portfolio(P)
        β := |10 / f(P_min)|                                              {starting gain value}
        T := M / 2
        R := 2 * N
        for t := 1 to T by +1 do
            for r := 1 to R by +1 do
                P_can := select_portfolio_randomly(P)
                for k := N to K+1 by −1 do
                    follow_Hopfield_dynamics(P_can)                       {P_can has k assets}
                    prune_worst_neuron(P_can)
                end for
                follow_Hopfield_dynamics(P_can)                           {P_can has K assets}
                evaluate_portfolio(P_can, H)                               {greedy algorithm}
                replace_maximum_portfolio(P_can, P)
            end for
            β := β / 0.95                                       {increasing gain values schedule}
        end for
    end for
    return H
end neural_network_heuristic
```

Algorithm 1. Neural network heuristic.

## 4. Computational experiments

In this section we present the results obtained when searching the general efficient frontier that solves the problem formulated in Eqs. (4)-(8). This efficient frontier has been computed using the former neural network (NN) and three other additional heuristic algorithms presented in [2] which are based on genetic algorithms (GA), tabu search (TS) and simulated annealing (SA), respectively. Regarding the execution time, we can say that there have not been any considerable differences between the neural network model and the other three heuristic methods.

We have done all the computational experiments with a set of benchmark data that has been already used in [2,3,4,7,11]. These data are from the Hang Seng index in Hong Kong and correspond to weekly prices from March 1992 to September 1997. The mean returns and covariances between these returns have been calculated for the data. This set of mean returns and covariances is publicly available at

*http://www.brunel.ac.uk/depts/ma/research/jeb/orlib/portinfo.html*

The test problem consists of $N=31$ different assets. Regarding all the results presented here, we have used the values $K=10$, $\varepsilon_i=0.01$ and $\delta_i=1$ for our problem formulation, and for the implementation of Algorithm 1 we have used the values $\Delta\lambda=0.1$ and $M=40$. So we have tested eleven values for the risk aversion parameter $\lambda$ and each one of the four heuristics has evaluated $M(N+1)=1280$ portfolios for each value of $\lambda$, without counting initialisations. As we have repeated it all three times with the purpose of reducing the effects of randomness, we count that each heuristic has evaluated a total of 42240 portfolios.

Taking the set of Pareto optimal portfolios obtained with each heuristic we can trace out four different heuristic efficient frontiers and compare them to the standard efficient frontier already shown in Fig. 1. Doing so we get an upper bound of the error associated to each heuristic algorithm. We show these comparisons in Fig. 2, where the standard efficient frontier is drawn in grey and the four heuristic efficient frontiers are drawn in black.

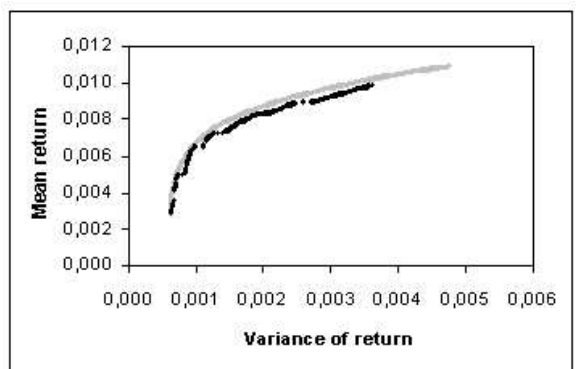 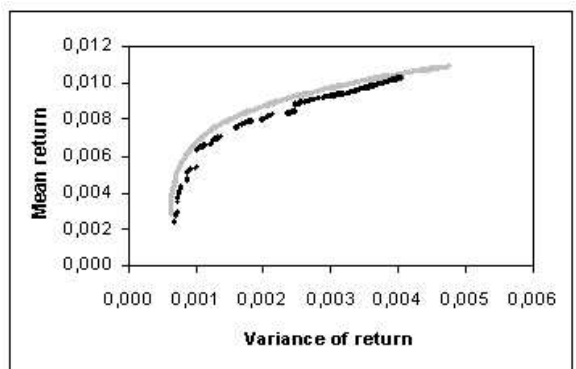

Fig. 2a. Genetic algorithm.    Fig. 2b. Tabu search.

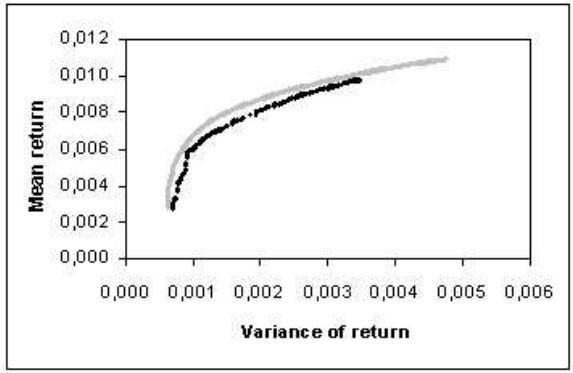 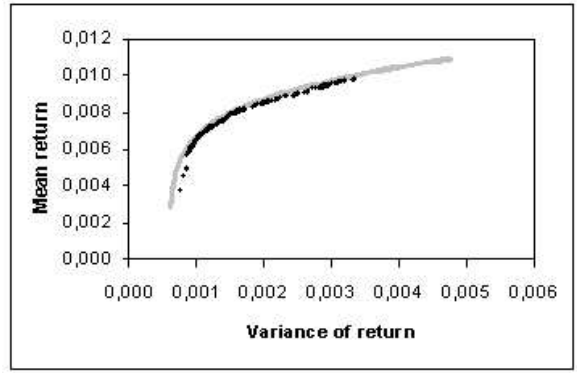

Fig. 2c. Simulated annealing.        Fig. 2d. Neural network.

Some numerical comparisons can also be done. First, in Table 1, we see how many of the 42240 portfolios evaluated by each heuristic method have persisted and finally appear in the corresponding heuristic efficient frontier. The results show the highest levels of persistence in the solutions from GA and TS, whilst the lowest level comes from NN.

Table 1
Persistence

| Heuristic | Cardinal | Percentage |
|---|---|---|
| GA | 466 | 1.10% |
| TS | 457 | 1.08% |
| SA | 392 | 0.93% |
| NN | 325 | 0.77% |

The above measurements only give an idea about the total number of solutions that appear in each heuristic efficient frontier, but they do not say anything about the quality of these solutions. Now we are going to define two distances, one for variances and another for mean returns. Let the pair $(v_i, r_i)$ represent the variance and mean return of a point in a heuristic efficient frontier. Let also $\hat{v}_i$ be the variance corresponding to $r_i$ according to a linear interpolation in the standard efficient frontier. We define the variance distance $\varphi_i$ from any heuristic point $(v_i, r_i)$ to the standard efficient frontier as the difference $\hat{v}_i - v_i$ (observe that this quantity will always be non negative). The average value of all these distances for the points in a heuristic efficient frontier gives us a distance between variances of the two efficient frontiers. In the same way, using the mean return $\hat{r}_i$ corresponding to $v_i$ according to a linear interpolation in the standard efficient frontier, we define the mean return distance $\psi_i$ as the difference $r_i - \hat{r}_i$. We get the average distance for the mean returns computing the average of all the distances $\psi_i$. Table 2 shows the values of both average distances. In both cases the best results are obtained with NN. GA gives the second best distances, whereas TS and SA share the last position in this classification.

Table 2
Average distance

| Heuristic | Mean return | Variance |
|---|---|---|
| GA | 0.000395 | 0.000270 |
| TS | 0.000587 | 0.000360 |
| SA | 0.000688 | 0.000313 |
| NN | 0.000212 | 0.000163 |

The inconvenience of the former metric is that a heuristic efficient frontier might be formed by a single point very close to the standard efficient frontier, giving two average distances better than those from any other heuristic efficient frontier with many more points lightly worse. For this reason we complete the average distances defining a pair of occupancy values. We take the interval of variance values covered by the standard efficient frontier and we divide it into 100 equal parts. Then, given all the points $(v_i, r_i)$ forming a heuristic efficient frontier, we compute the percentage of subintervals that have any variance $v_i$ projected onto them. This quantity gives us and idea of the occupancy level for the 100 variance subintervals. Dividing the interval of mean return values into 100 equal parts and counting the number of subintervals that have any mean return $r_i$ projected onto them, we get the occupancy level for the mean return subintervals. Both results are shown in Table 3. The heuristic efficient frontiers corresponding to GA and SA cover more subintervals than the other two, specially the NN efficient frontier.

Table 3
Occupancy

| Heuristic | Mean return | Variance |
|---|---|---|
| GA | 74% | 67% |
| TS | 59% | 67% |
| SA | 75% | 66% |
| NN | 55% | 56% |

In order to improve these results obtained separately by the four heuristic algorithms, we merge the four heuristic efficient frontiers into a single one and we remove from it the dominated solutions. The resulting merged efficient frontier can be seen in Fig. 3.

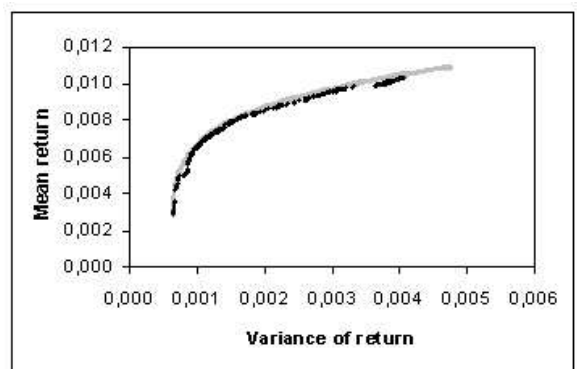

Fig. 3. Merged efficient frontier.

If we now separate this merged efficient frontier into the four parts that form it according to the heuristic origin of the points, we get the result shown in Fig. 4.

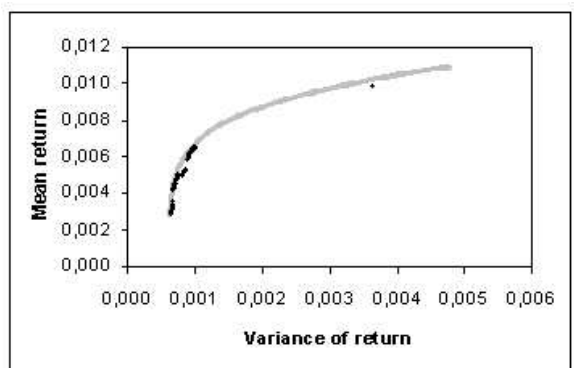 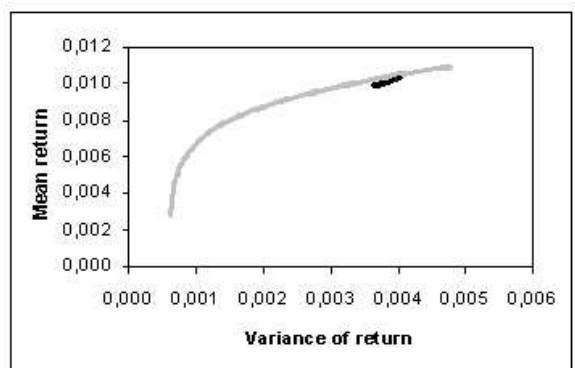

Fig. 4a. Genetic algorithm.   Fig. 4b. Tabu search.

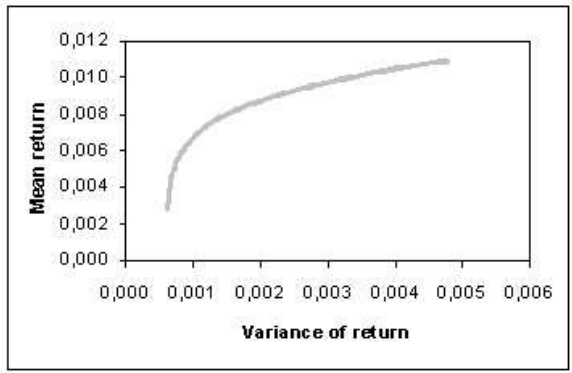 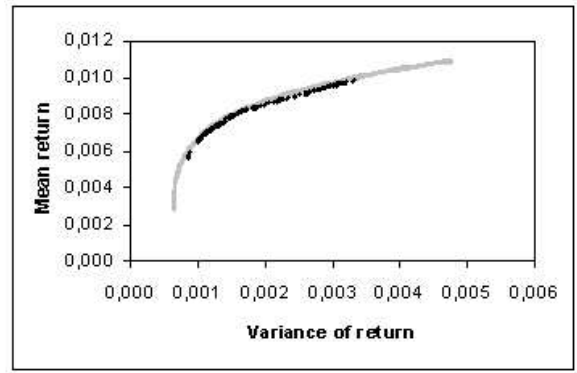

Fig. 4c. Simulated annealing.                    Fig. 4d. Neural network.

Let us now take a look at this new efficient frontier. First we use it to compare the quality of the initial solutions provided by each heuristic method. We take the initial number of points in each heuristic efficient frontier (Table 1) and we compare it with the final number of points in the merged efficient frontier that come from the respective heuristic. Table 4 shows the percentage of points surviving the merge process. The quality of NN solutions is outstanding. On the contrary, all SA solutions are dominated by solutions from the other three heuristic methods.

Table 4
Points surviving the merge process

| Heuristic | Initial cardinal | Final cardinal | Percentage |
|---|---|---|---|
| GA | 466 | 105 | 23% |
| TS | 457 | 111 | 24% |
| SA | 392 | 0 | 0% |
| NN | 325 | 284 | 87% |

Next we compare in Table 5 the number of points in which each heuristic contributes to the merged efficient frontier (in our test results this efficient frontier had exactly 500 different points). More than half of the points in the merged efficient frontier come from the NN. This result highlights again the quality of the NN solutions.

Table 5
Contribution to the merge process

| Heuristic | Cardinal | Percentage |
|---|---|---|
| GA | 105 | 21% |
| TS | 111 | 22% |
| SA | 0 | 0% |
| NN | 284 | 57% |

Let us now repeat the computation of the average distances $\varphi_i$ and $\psi_i$. Table 6 shows the new results for the merged efficient frontier as a whole, as well as for each one of its four parts separately. When we consider mean returns, we get the best results with NN. But when we consider variances, GA gives the best solutions. It must also be noticed that the solutions from NN are better in average terms, since they are the second best regarding variances whilst GA solutions are the worst ones regarding mean returns.

Table 6
Average distance after the merge process

| Heuristic | Mean return | Variance |
|---|---|---|
| All | 0.000214 | 0.000142 |
| GA | 0.000311 | 0.000036 |
| TS | 0.000276 | 0.000391 |
| SA | – | – |
| NN | 0.000154 | 0.000084 |

Finally, in Table 7 we can see the occupancy results obtained with the merged efficient frontier. The best occupancy levels are those from NN. This result also confirms the high quality of NN solutions. The second best occupancy levels are those from GA.

Table 7
Occupancy after the merge process

| Heuristic | Mean return | Variance |
|---|---|---|
| All | 80% | 70% |
| GA | 32% | 10% |
| TS | 6% | 11% |
| SA | 0% | 0% |
| NN | 46% | 53% |

## 5. Conclusions

In this work we have focused on solving the portfolio selection problem and tracing out its efficient frontier. Instead of using the standard Markowitz mean-variance model, we have used a generalization of it that includes cardinality and bounding constraints. Dealing with this kind of constraints, the portfolio selection problem becomes a mixed quadratic and integer programming problem for which no computational efficient algorithms are known.

We have developed a heuristic method based on the Hopfield neural network and we have used it to solve the general mean-variance portfolio selection model. The results obtained have been compared to those obtained using three other heuristic methods coming from the fields of genetic algorithms, tabu search and simulated annealing.

All the experimental results presented in this paper lead us to conclude that none of the four heuristic methods has outperformed the others in all the comparisons considered. Anyway, we must specially mention the fact that the neural network model has given us a set of solutions with higher quality than the solutions from the other three heuristic methods, although they did not outstand in their number.

### Acknowledgements

This work was supported in part by CICYT projects FPA2002-04452-C02-02 and FPA2002-04208-C07-07. A.F. also thanks Universitat Rovira i Virgili for financial support.